# Roof Damage Assessment from Automated 3D Building Models

Kenichi Sugihara[1], Martin Wallace[2*], Kongwen (Frank) Zhang[3], Youry Khmelevsky[4]

1. Doctor of Engineering, Professor, Gifu-Keizai University, 5-50 Kitagata-chou Ogaki-city Gifu-Pref. 503-8550, Japan, +81 584-77-3511, mjsbp812@yahoo.co.jp

2. Undergraduate Student, Okanagan College, 1000 KLO Rd., Kelowna, BC V1Y 4X8, Canada, +1 250-801-7976, martin.v.wallace@ieee.or

3. Lecturer, Selkirk College, 301 Frank Beinder Way, Castlegar, BC V1N 4L3, Canada, +1 250-304-6527, fzhang@selkirk.ca

4. Professor, Okanagan College, 1000 KLO Rd., Kelowna, BC V1Y 4X8, Canada, +1 250-762-5445, ykhmelevsky@okanagan.bc.ca

* Martin Wallace: Undergraduate Student, Okanagan College, martin.v.wallace@ieee.org

## ABSTRACT

The 3D building modeling is important in urban planning and related domains that draw upon the content of 3D models of urban scenes. Such 3D models can be used to visualize city images at multiple scales from individual buildings to entire cities prior to and after a change has occurred. This ability is of great importance in day-to-day work and special projects undertaken by planners, geo-designers, and architects. In this research, we implemented a novel approach to 3D building models for such matter, which included the integration of geographic information systems (GIS) and 3D Computer Graphics (3DCG) components that generates 3D house models from building footprints (polygons), and the automated generation of simple and complex roof geometries for rapid roof area damage reporting. These polygons (footprints) are usually orthogonal. A complicated orthogonal polygon can be partitioned into a set of rectangles. The proposed GIS and 3DCG integrated system partitions orthogonal building polygons into a set of rectangles and places rectangular roofs and box-shaped building bodies on these rectangles. Since technicians are drawing these polygons manually with digitizers, depending on aerial photos, not all building polygons are precisely orthogonal. But, when placing a set of boxes as building bodies for creating the buildings, there may be gaps or overlaps between these boxes if building polygons are not precisely orthogonal. In our proposal, after approximately orthogonal building polygons are partitioned and rectified into a set of mutually orthogonal rectangles, each rectangle knows which rectangle is adjacent to and which edge of the rectangle is adjacent to, which will avoid unwanted intersection of windows and doors when building bodies combined.

## 1. INTRODUCTION

A 3D city model, such as the one shown in Fig. 1 right, is important in urban planning and gaming industries. Urban planners may draw the maps for sustainable development. 3D city models based on these maps are quite effective in understanding what if this alternative plan is realized. However, enormous time and labor has to be consumed to create these 3D models, using 3D modelling software such as 3ds Max or SketchUp. In order to automate laborious steps, we proposed a GIS (Geographic Information System) and CG integrated system (Sugihara et al. 2005, 2006, 2012) for automatically generating 3D building models, based on only building polygons i.e. building footprints on a digital map shown in Fig. 1 left.

A complicated orthogonal polygon can be partitioned into a set of rectangles. The proposed integrated system partitions orthogonal building polygons into a set of rectangles and places rectangular roofs and box-shaped building bodies on these rectangles. In order to partition an orthogonal polygon, we also proposed a useful polygon expression (RL expression: edges' Right & Left turns expression) and a partitioning scheme for deciding from which vertex a dividing line (DL) is drawn (Sugihara et al. 2012).

Since technicians are drawing building polygons manually with digitizers, depending on aerial photos or satellite imagery as shown in Fig. 1 left, not all building polygons are precisely orthogonal. When placing a set of boxes as

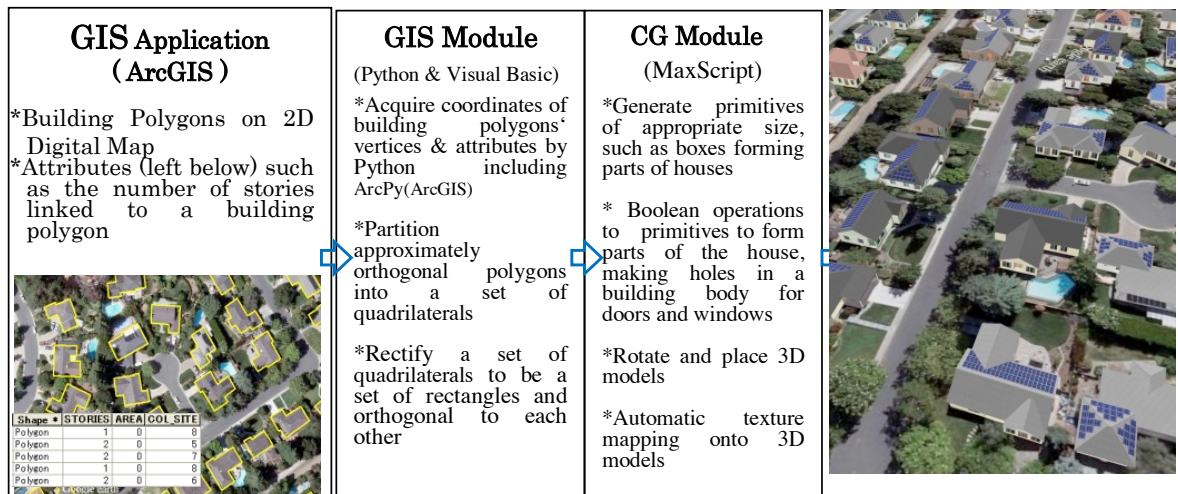

Automatically generated 3D urban

**Fig. 1: Pipeline of Automatic Generation for 3D building Models**

building bodies for forming the buildings, there may be gaps or overlaps between these boxes if building polygons are not strictly orthogonal. Our contribution is the new methodology for rectifying the shape of building polygons and constructing 3D building models without any gap and overlap. In our proposal, after approximately orthogonal building polygons are partitioned and rectified into a set of mutually orthogonal rectangles, each rectangle knows which rectangle is adjacent to and which edge of the rectangle is adjacent to, which will avoid unwanted intersect ion of windows and doors when building bodies combined.

## 2. RELATED WORK

Since 3D urban models are important information infrastructure that can be utilized in several fields, the researches on creations of 3D urban models are in full swing. Various types of technologies, ranging from computer vision, computer graphics, photogrammetry, and remote sensing, have been proposed and developed for creating 3D urban models. Procedural modelling is an effective technique to create 3D models from sets of rules such as L-systems, fractals, and generative modelling language (Parish et al. 2001). Müller et al. (2006) have created an archaeological site of Pompeii and a suburbia model of Beverly Hills by using a shape grammar that provides a computational approach to the generation of designs. They import data from a GIS database and try to classify imported mass models as basic shapes in their shape vocabulary. If this is not possible, they use a general extruded footprint together with a general roof obtained by the straight skeleton computation defined by a continuous shrinking process (Aichholzer et al. 1995).

By using the straight skeleton, Kelly et al. (2011) present a user interface for the exterior of architectural models to interactively specify procedural extrusions, a sweep plane algorithm to compute a two-manifold architectural surface.

More recently, image-based capturing and rendering techniques, together with procedural modelling approaches, have been developed that allow buildings to be quickly generated and rendered realistically at interactive rates. Bekins et al. (2005) exploit building features taken from real-world capture scenes. Their interactive system subdivides and groups the features into feature regions that can be rearranged to texture a new model in the style of the original. The redundancy found in architecture is used to derive procedural rules describing the organization of the original building, which can then be used to automate the subdivision and texturing of a new building. This redundancy can also be used to automatically fill occluded and poorly sampled areas of the image set.

Aliaga et al. (2007) extend the technique to inverse procedural modelling of buildings and they describe how to use an extracted repertoire of building grammars to facilitate the visualization and modification of architectural structures. They present an interactive system that enables both creating new buildings in the style of others and modifying existing buildings in a quick manner.

Vanega et al. (2010) interactively reconstruct 3D building models with the grammar for representing changes in building geometry that approximately follow the Manhattan-world (MW) assumption which states there is a predominance of three mutually orthogonal directions in the scene. They say automatic approaches using laser-scans or LIDAR data, combined with aerial imagery or ground-level images, suffering from one or all of low-resolution sampling, robustness, and missing surfaces. One way to improve quality or automation is to incorporate assumptions about the buildings such as MW assumption.

Jianxiong (2014) presents a 3D reconstruction and visualization system to automatically produce clean and well-regularized texture-mapped 3D models for large indoor scenes, from ground-level photographs and 3D laser points. The key component is a new algorithm called "Inverse CSG" for reconstructing a scene in a Constructive Solid Geometry (CSG) representation consisting of volumetric primitives, which imposes regularization constraints to exploit structural regularities. However, with the lack of ground-truth data preventing them from conducting quantitative reconstruction accuracy evaluations, they have to manually overlay their model with a floor plan image.

By these interactive modelling, 3D building models with plausible detailed façade can be achieved. However, the limitation of these modelling is the large amount of user interaction involved (Nianjuan et al. 2009). When creating 3D urban models for urban planning or facilitating public involvement, 3D urban models should cover lots of citizens' and stakeholders' buildings involved. This means that it will take an enormous time and labor to model a 3D urban model with hundreds of buildings.

Thus, the GIS and CG integrated system that automatically generates 3D urban models immediately is proposed, and the generated 3D building models that constitute 3D urban models are approximate geometric 3D building models that citizens and stakeholder can recognize as their future residence or real-world building.

## 3. PIPELINE of AUTOMATIC GENERATION

As shown in Fig. 1, the proposed automatic building generation system consists of GIS application (ArcGIS, ESRI Inc.), GIS module and CG module. The source of the 3D urban model is a digital residential map that contains building polygons linked with attributes data shown in Fig. 1 left bellow, consist of the number of stories, the image code of roof, wall and the type of roof (gable roof, hipped roof, gambrel roof, mansard roof, temple roof and so forth). If a digital map is given, the only manual labor is to input these attributes. The maps are then preprocessed at the GIS module, and the CG module finally generates the 3D urban model. As a GIS module, a Python program including ArcPy(ArcGIS) acquires coordinates of building polygons' vertices and attributes. Preprocessing at the GIS module includes the procedures as follows:

**(1)** Filter out an unnecessary vertex whose internal angle is almost 180 degrees.
**(2)** Partition or separate approximately orthogonal polygons into a set of quadrilaterals.
**(3)** Generate inside contours by straight skeleton computation for placing doors, windows, fences and shop façades which are setback from the original building polygon.
**(4)** Rectify a set of quadrilaterals to be a set of rectangles and orthogonal to each other.
**(5)** Export the coordinates of polygons' vertices, the length, width and height of the partitioned rectangle, and attributes of buildings.

The CG module receives the pre-processed data that the GIS module exports, generating 3D building models. In GIS module, the system measures the length and inclination of the edges of the partitioned rectangle. The CG module generates a box of the length and width, measured in GIS module. In case of modelling a building with roofs, the CG module follows these steps:

**(1)** Generate primitives of appropriate size, such as boxes, prisms or polyhedra that will form the various parts of the house.
**(2)** Boolean operations applied to these primitives to form the shapes of parts of the house, for examples, making holes in a building body for doors and windows, making trapezoidal roof boards for a hipped roof and a temple roof.
**(3)** Rotate parts of the house according to the inclination of the partitioned rectangle.
**(4)** Place parts of the house.
**(5)** Texture mapping onto these parts according to the attribute received.
**(6)** Copy the 2nd floor to form the 3rd floor or more in case of building higher than 3 stories.

CG module has been developed using Maxscript that controls 3D CG software (3ds MAX, Autodesk Inc).

## 4. FUNCTIONALITY OF GIS MODULE

### 4.1 Polygon Expression and Partitioning Scheme

At map production companies, technicians are drawing building polygons manually with digitizers, depending on aerial photos or satellite imagery as shown in Fig. 1 and 2. This aerial photo and digital map also show that most building polygons are approximately orthogonal polygons. An approximately orthogonal polygon can be replaced by a combination of rectangles. When the vertices of a polygon are numbered in clockwise order and edges of a polygon are followed clockwise, an edge turns to the right or to the left by 90 degrees. It is possible to assume that an orthogonal polygon can be expressed as a set of its edges' turning direction; an edge turning to the 'Right' or to the 'Left'.

We proposed a useful polygon expression (RL expression: edges' Right and Left turns expression) for specifying the shape pattern of an orthogonal polygon. For example, an orthogonal polygon with 22 vertices shown in Fig. 2 is expressed as a set of its edges' turning direction; RRLRLRLRLRRRLLRRLRLRLR where R and L mean a change of an edge's direction to the right and to the left, respectively.

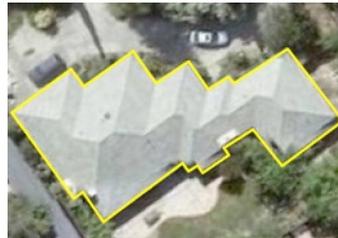
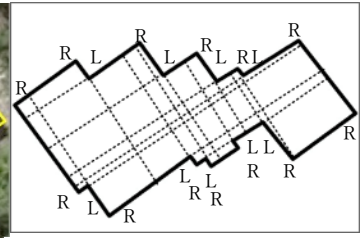

Fig.2: Building polygons on satellite image, expressed as a set of its edges' turning direction; RRLRLRLRLRRRLLRRLRLRLR

Fig.3: Two DLs from L vertex. 9 L vertices, so 18 DLs can be drawn from each L vertex

The more vertices a polygon has, the more partitioning scheme a polygon has, since the interior angle of a 'L' vertex is 270 degrees and two DLs (dividing lines) can be drawn from a L vertex. In the polygon shown in Fig. 3, there are 9 L vertices, so 18 DLs can be drawn from each L vertex as shown in dotted lines. In our proposal, among many possible DLs, the DL that satisfies the following conditions is selected for partitioning:

(1) *A DL that cuts off 'one rectangle'.*
(2) *A DL whose length is shorter than the width of a 'main roof' that a 'branch roof' is supposed to extend to, where a 'branch roof' is a roof that is cut off by a DL and extends to a main roof.*

How the system is finding 'branches' is as follows. The system counts the number of consecutive 'R' vertices (=$n_R$) between 'L' vertices. If $n_R$ is two or more, then it can be a branch. One or two DLs can be drawn from 'L' vertex in a clockwise or counter-clockwise direction, depending on the length of the adjacent edges of 'L' vertex. How the DL is drawn (clockwise or counter-clockwise), that is, 'dividing pattern' is used for reconstructing a rectified polygon and saved at the divided rectangle.

### 4.2 Process of Polygon Partition and Shape Rectification

Fig. 4 shows detailed process of polygon partition and shape rectification, generation of a 3D building model. The system executes the partitioning procedure as follows:

(1) Classify the branches by the number of successive 'R' vertices, and the length of the edge especially incident to 'L' vertex and dividing pattern.
(2) Check whether the DL is satisfying three conditions or not, by measuring the distance between the DL and other edges in the same polygon, which will be the width of the main roof.
(3) If the DL satisfies the conditions and is given the highest priority, then the position of the intersection between the DL and edges is calculated.
(4) Set the erase flags for the vertices of the branch that are cut off from a body polygon, and a new vertex that is the intersection will be included by the body polygon.
(5) Measure the edge length and inner angle of the polygon's vertices and acquire edges' Right and Left turns expression (RL expression).

This partitioning procedure continues until the number of the vertices of the body polygon is four. In Fig. 4 (2) in upper row, branch quad is divided by DL1 which is a Forward Dividing Line (FDL) in terms of polygon vertices numbering (clockwise). In Fig. 4 (3) in upper & middle row, branch quad is divided by DL2 which is a Backward Dividing Line (BDL) drawn in the opposite direction (counter clockwise). The dividing pattern is defined by this FDL or BDL.

A branch quad divided by DL1 in Fig. 4 (2) in lower row is given the highest priority for partitioning, since this partition reduces the number of the vertices of the body polygon by four. A branch quad divided by DL1 in Fig. 4 (2) in upper & middle row are given the less priority for partitioning, as this partition reduces the vertex number by two. Thus, the system is giving each DL the degree of priority for partitioning, and the partitioning by the DL of the highest priority will be executed. Two DLs satisfying first two conditions can be drawn from the same vertex, such as 'A', shown in Fig. 4 (2) in lower row. But, the third condition excludes the longer DL, since condition third demands a DL whose vertices are not shared by another DL, which will prevent unnecessary dividing. A shorter DL is selected for partitioning. After remaining body polygon's vertex number is four, the shape rectification begins by transforming the remaining polygon into a rectangle, as shown in Fig. 4 (5). An active branch quad will start to search for an adjacent quad in reverse order from the last divided branch. The active quad will find a neighboring quad by using quad's vertices position of before rectification. Therefore, each quad instance has double vertices positions of before and after rectification. When forming the branch rectangle and branch roof rectangle, the system is using quad's vertices position of after rectification.

## 4.3 SHAPE RECTIFICATION

Specifically, the rectification procedure is implemented to the polygon shown in Fig. 5, which shows the process of polygon partition and shape rectification, automatic 3D modelling. Before polygon partitioning, all edge length and edge inclination of the polygon are measured, and the length of all edges are sum up according to the snapped angle of all edge inclination. Then, the angle for a longest sum up edge length can be adopted as the 'main angle' of the polygon, which will be then used as the inclination of all partitioned rectangles. After GIS module measuring the length and inclination of all edge of the partitioned polygon, i.e., a quadrilateral ('quad' for short), the edges are categorized into a long edge (w_L or edge12) and a short edge (w_S or edge23). A partitioned quad (quadrilateral) is numbered clockwise with the start point of a longest edge facing right as pt1 (a1, b1,..) or with the start point of a longest edge facing left as pt3 (a3, b3,..) as shown in Fig. 5 (a).

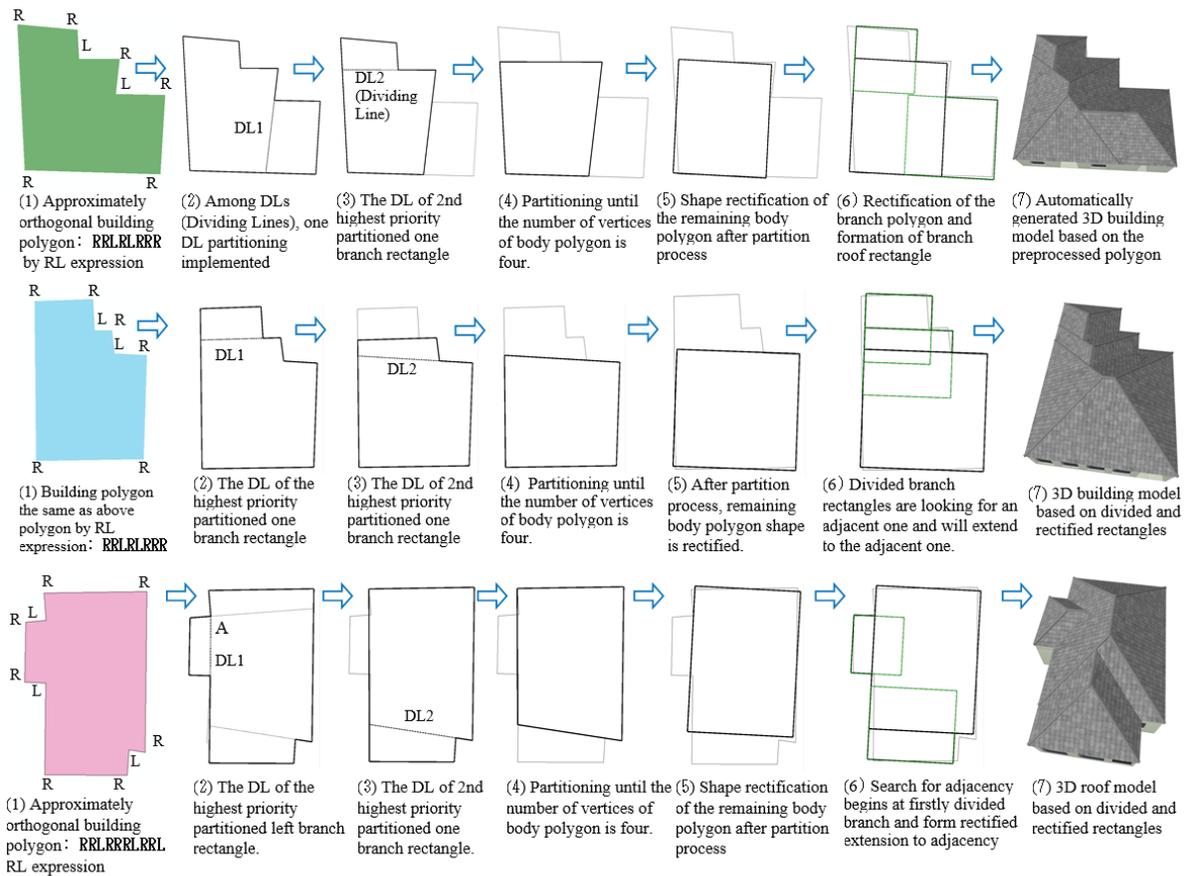

Fig.4: Detailed process of polygon partition and shape rectification, generation of 3D model

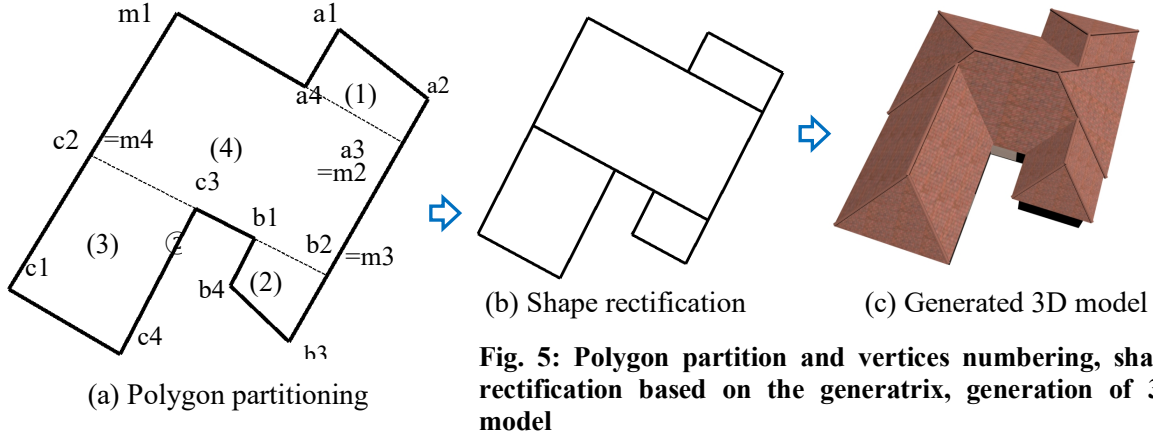

(a) Polygon partitioning  (b) Shape rectification  (c) Generated 3D model

**Fig. 5:** Polygon partition and vertices numbering, shape rectification based on the generatrix, generation of 3D model

When a quad is cut off, the dividing pattern and which edge of the quad is cut off (active edge i.e. DL: Dividing Line) is saved at the quad. During the searching stage, an active quad will search for an adjacent quad by locating which quad the checking point on the active edge contains, and then checking on which edge of the adjacent quad the checking point is. In case of quad (1) in Fig. 5 (a), DL (a3a4) will be an active edge, and search for an adjacent quad. After searching and having found out the adjacent quad is quad (4) and the adjacent edge is m1m2 of quad (4), the mutual vertex is a3(=m2), which the rectification procedure uses as a 'standard position (generatrix)' for rectification, since this vertex is shared by two quads and could be an origin of local coordinates. The rectified positions of the vertices of quad (1) are calculated as follows.

$a1.x = m2.x + w\_S*\cos\theta - w\_L*\sin\theta$
$a1.y = m2.y + w\_S*\sin\theta + w\_L*\cos\theta$
$a2.x = m2.x + w\_S*\cos\theta : a2.y = m2.y + w\_S*\sin\theta$
$a4.x = m2.x - w\_L*\sin\theta : a4.y = m2.y + w\_L*\cos\theta$

where $\theta$ is the main angle and w_S is the average length of two short sides of the rectangle, and w_L is the average length of two long sides of the rectangle. In case of quad (3), the mutual vertex is c2(=m4), which the rectification procedure also uses as a standard position for rectification. The rectified positions of the vertices of quad (3) are calculated as follows.

$c1.x = m4.x - w\_L*\cos\theta : c1.y = m4.y - w\_L*\sin\theta$
$c3.x = m4.x + w\_S*\sin\theta : c3.y = m4.y - w\_S*\cos\theta$
$c4.x = m4.x - w\_L*\cos\theta + w\_S*\sin\theta$
$c4.y = m4.y - w\_L*\sin\theta - w\_S*\cos\theta$

The rectified positions of the vertices of a branch quad in other cases are calculated likewise according to the dividing pattern and which edge of the branch quad is cut off.

## 5. FUNCTIONALITY OF CG MODULE

As shown in Fig. 1, the CG module receives the pre-processed data that the GIS module exports, generating 3D building models. The pre-processed data includes the following data; the number of partitioned rectified quads, vertex coordinates of rectified building body quads and rectified roof quads, lengths and inclinations of the long and short edge of the quads (w_L, w_S), attributes linked with building polygon consisting of the number of stories, the image code of roof, wall and the type of roof (flat, gable roof, hipped roof, oblong gable roof, gambrel roof, mansard roof, temple roof and so forth). Depending on these data, a CG module generates 3D building models. Fig. 6 shows the generation process of a hipped roof house model by CG module.

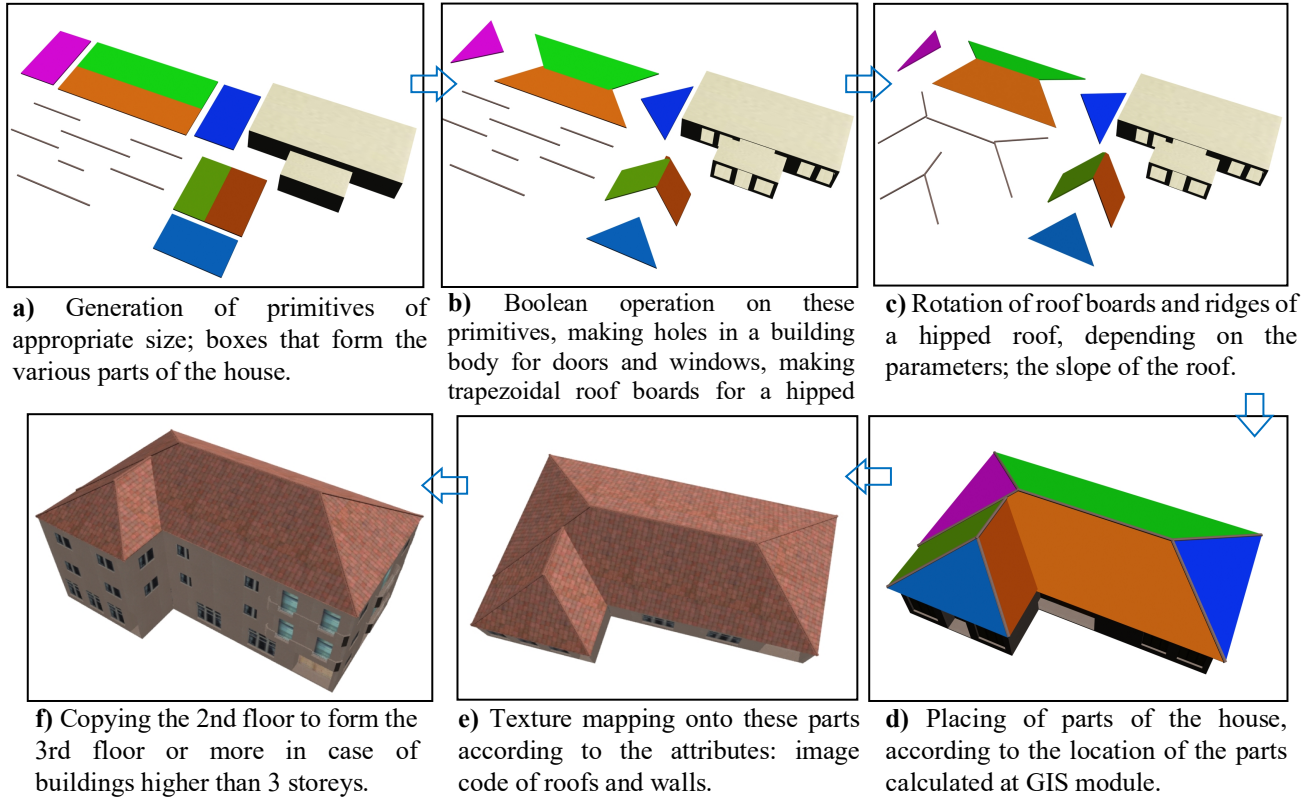

**a)** Generation of primitives of appropriate size; boxes that form the various parts of the house.

**b)** Boolean operation on these primitives, making holes in a building body for doors and windows, making trapezoidal roof boards for a hipped

**c)** Rotation of roof boards and ridges of a hipped roof, depending on the parameters; the slope of the roof.

**f)** Copying the 2nd floor to form the 3rd floor or more in case of buildings higher than 3 storeys.

**e)** Texture mapping onto these parts according to the attributes: image code of roofs and walls.

**d)** Placing of parts of the house, according to the location of the parts calculated at GIS module.

**Fig. 6: Generation process of a 3D building model in CG module**

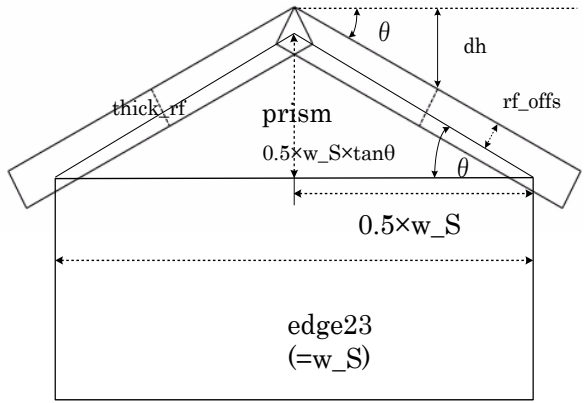

The width of a roof board is as follows.
$wid\_rfb = side23L + eaves23 + rf\_offs \times \tan\theta$
Here, $side23L = 0.5 \times w\_S \times \sqrt{1+\tan^2\theta}$

The height of a roof board is as follows.
$hei\_rf = st\_heit - 0.5 \times (side23L + eaves23 + rf\_offs \times \tan\theta) \times$
$- thick\_rf \times \cos\theta + rf\_offs/\cos\theta + 0.5 \times \tan\theta \times w\_S$

Here, 'st_heit' is start height as follows.
$st\_heit =$ (floor-to-floor height)×(the number of stor

**Fig. 7: Front view of a gable roof**

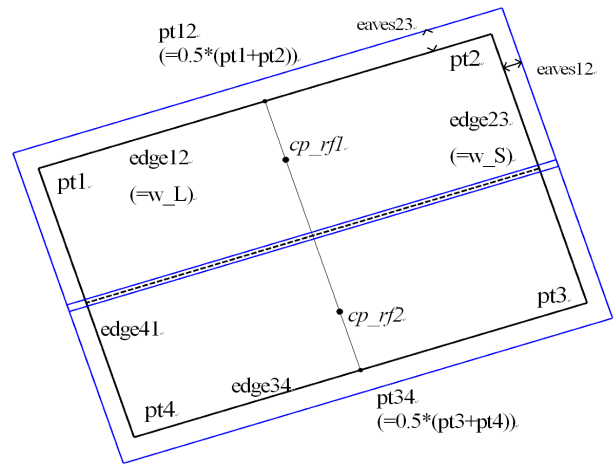

$ratio\_s = 0.25 - \dfrac{0.5 \times (eaves23 \times \cos\theta + rf\_offs * \sin\theta)}{w\_S} + \dfrac{thick\_rf \times \sin\theta}{w\_S}$

$cp\_rf1 = (1.0 - ratio\_s) \times pt12 + ratio\_s \times pt34$
$cp\_rf2 = ratio\_s \times pt12 + (1.0 - ratio\_s) \times pt34$

Here, 'thick_rf' is a thickness of a roof board. 'eaves23' is the length of eaves in a direction of edge23. $\theta$ is an angle of a roof slope with a horizontal plane. 'rf_offs' is the offset of a roof board from a prism as shown Figure 12.

**Fig. 8: Ground plan and parameters of a gable roof**

As mentioned earlier, the GIS module partitions and rectifies approximately orthogonal building polygons into a set of rectangles, and the CG module places rectangular roofs and box-shaped building bodies on these rectangles, depending on the front view and ground plan such as Fig. 7 and Fig. 8. The vertices of the rectangle are numbered clockwise with the upper left vertex of a long edge being numbered 'pt1' as shown in Fig. 8, 'ground plan' of a gable roof. The length of edge12 and edge23 are w_L and w_S respectively.

In 3ds Max used for the creation of 3D models, each building part or primitive has its own control point ('cp') and local coordinates that control its position and direction. The position of a 'cp' is different in each primitive. As shown in Fig. 8, the top of a gable roof consists of two roof boards (two thin boxes). Since the 'cp' of a box lies in a center of a base, it is placed on the point that divides the line through pt12 and pt34 at the ratio shown in ground plan. The height of the 'cp's of two roof boards is shown in the front view of a gable roof (Fig. 7).

The CG module's generation process for modeling a hipped roof house is as follows:

**(1)** Generation of primitives of appropriate size, such as boxes, prisms that are the parts of the house: For example, the length and width of a box as a house body are decided by the rectangle partitioned and rectified at GIS module. As shown in Fig. 8, the length of a box will be w_L and the width of a box will be w_S. Also, the length of a thin box as a roof board is decided by the rectangle partitioned while the width of a roof board is decided by the slope of the roof given as a parameter, as shown right in Fig. 7.
**(2)** Boolean operation on these primitives to form the shapes of parts of the house: for examples, making holes in a house body for doors and windows, making trapezoidal or triangular roof boards for a hipped roof. The size and position, number of the holes are decided by the given parameters.
**(3)** Rotation of parts of the house: Parts of the house are rotated depending on the direction of the rectangle partitioned. The roof boards are rotated so as to align them according to, respectively, the slopes of the roofs.
**(4)** Positioning of parts of the house: Parts of the house are placed depending on the position of the rectangle partitioned. For example, the control points of roof boards are placed as shown in Fig. 7 and Fig. 8.
**(5)** Texture mapping onto these parts according to the attribute data, such as image code of wall and roof, stored and administrated at GIS application.
**(6)** Copying the 2nd floor to form the 3rd floor or more in case of building higher than 3 stories according to the attribute data, the number of stories as shown in Fig. 6 (f).

## 6. EXAMPLES and CONCLUSIONS

Here are the examples of automatic generation process of 3D building models and roof damage assessment report in Fig. 9. We proposed and presented the automated generation of simple and complex roof geometries for rapid roof area damage reporting. The source of complex roof geometries is the footprints (building polygons) which are usually orthogonal. A complicated orthogonal polygon can be partitioned into a set of rectangles. The proposed GIS and 3DCG integrated system partitions orthogonal building polygons into a set of rectangles and places rectangular roofs and box-shaped building bodies on these rectangles. Since these polygons are manually drawn, depending on aerial photos, not all building polygons are precisely orthogonal. But, when placing a set of boxes as building bodies for creating the buildings, there may be gaps or overlaps between these boxes if building polygons are not precisely orthogonal. In our proposal, after approximately orthogonal building polygons are partitioned and rectified into a set of mutually orthogonal rectangles, each rectangle knows which rectangle is adjacent to and which edge of the rectangle is adjacent to, which will avoid unwanted intersection of windows and doors when building bodies combined for automatic generation of 3D building models.

Our future work includes extension of geometric primitive types to more shapes, such as elliptic cylinders or spheres.

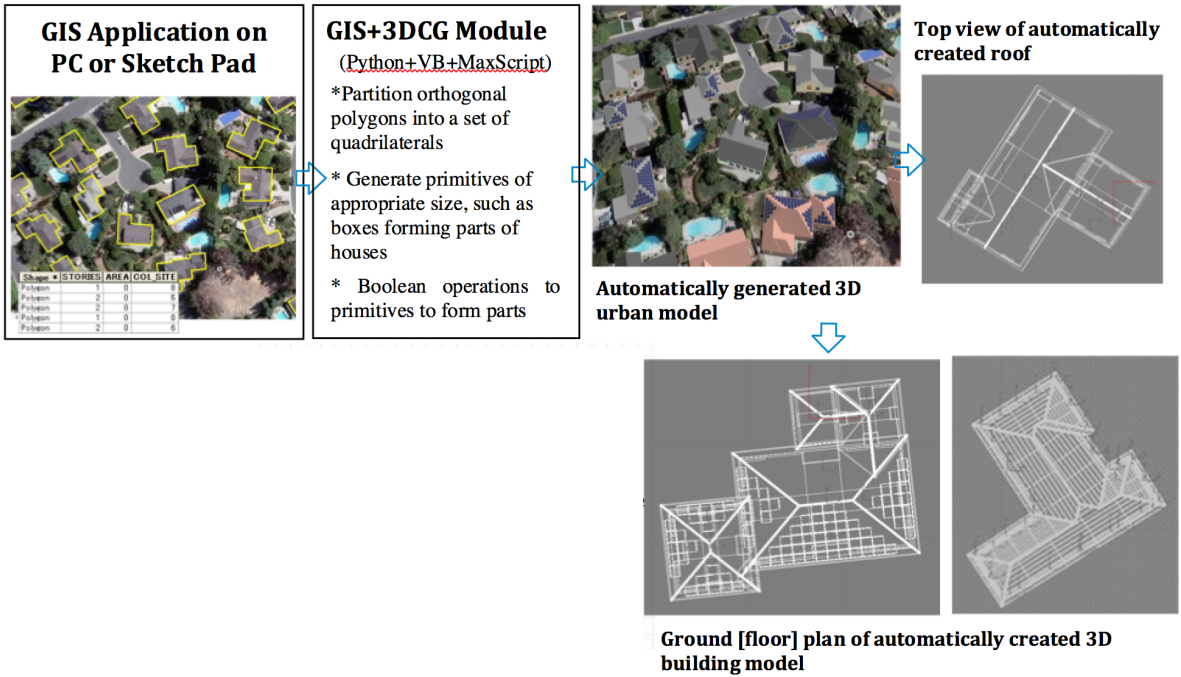

**Fig. 9: Automatic generation process of 3D building models and Roof Damage Assessment Report**